\documentclass[letterpaper, 10 pt, conference]{ieeeconf}
\usepackage{amsmath}
\usepackage{amssymb}   

\usepackage{algorithm}
\usepackage{algorithmic}
\usepackage{float}
\usepackage[caption=false, font=footnotesize]{subfig}
\usepackage{graphicx}
\newtheorem{assumption}{Assumption}

\DeclareMathOperator*{\argmin}{arg\,min}
\DeclareMathOperator*{\argmax}{arg\,max}

\IEEEoverridecommandlockouts
\title{\LARGE \bf
    Distributed Multi-Robot Multi-Target Simultaneous Search and Tracking in an Unknown Non-convex Environment
}

\author{Jun Chen, Jiaqing Ma, and Philip Dames% <-this % stops a space
\thanks{J.~Chen and J.~Ma are with the School of Electrical and Automation Engineering, Nanjing Normal University, Nanjing, Jiangsu 210023, China}
\thanks{P.~Dames is with the Department of Mechanical Engineering, Temple University, Philadelphia, PA 19122, USA}%
}

\begin{document}

\maketitle

\begin{abstract}
In unknown non-convex environments, such as indoor and underground spaces, deploying a fleet of robots to explore the surroundings while simultaneously searching for and tracking targets of interest to maintain high-precision data collection represents a fundamental challenge that urgently requires resolution in applications such as environmental monitoring and rescue operations.
Current research has made significant progress in addressing environmental exploration, information search, and target tracking problems, but has yet to establish a framework for simultaneously optimizing these tasks in complex environments.
In this paper, we propose a novel motion planning algorithm framework that integrates three control strategies: a frontier-based exploration strategy, a guaranteed coverage strategy based on Lloyd's algorithm, and a sensor-based multi-target tracking strategy. 
By incorporating these three strategies, the proposed algorithm balances coverage search and high-precision active tracking during exploration.
Our approach is validated through a series of MATLAB simulations, demonstrating validity and superiority over standard approaches.
\end{abstract}

\begin{keywords}
    Multi-Robot Systems, Active Sensing, Distributed Systems.
\end{keywords}

\section{Introduction}
Simultaneously searching for and tracking multiple unknown targets constitutes a fundamental challenge in multi-robot collaboration, with applications spanning search and rescue to environmental monitoring.
In these tasks, a robot team must collaboratively estimate the target state with unknown or partially known prior information and execute motion planning to achieve environmental coverage for undetected target search, while simultaneously performing information-rich path planning to ensure precise tracking of detected targets.
While multi-target search and tracking for multi-robot systems (MRSs) is well-established in known environments, deploying such systems in unknown, non-convex environments remains a practical challenge.
This stems from the dual requirement to efficiently explore unstructured, unknown task spaces while ensuring the accuracy of online target information gathering.

\subsection{Multi-Robot Exploration}
Multi-robot cooperative exploration algorithms are a prerequisite for active sensing in unknown or partially unknown maps, including behavior-based methods, sample-based methods, learning-based methods, and frontier-based methods.
The use of behavior-based methods, such as particle swarm optimization (PSO) \cite{couceiro2011novel}, generate simple local control rules based on biological societies (e.g., ants, bees, birds) to develop similar behaviors in distributed cooperative MRS, including exploration of unknown environments.
The next-best-view (NBV) method \cite{gonzalez2002navigation} exemplifies a sample-based method, where a finite set of randomly generated candidate positions is evaluated to efficiently select the next best view.
Learning-based approaches \cite{yu2023asynchronous} model the entire exploration process by training a single neural network without prior knowledge to find a flexible strategy from RL policies via interaction with environments. 
Among all strategies for multi-robot exploration, frontier-based approaches \cite{yamauchi1997frontier, burgard2005coordinated} are the most widely applied, which involves robots continuously identifying frontiers, i.e., edges where explored space meets unexplored regions. 
These frontiers serve as candidate targets, guiding robots toward the zones that promise the highest information gain. 
The frontier-based method can generate targets directly from the frontier cluster based on indicators such as the shortest Euclidean distance, thereby achieving extremely fast computation speed \cite{wang2025multi}.

There is limited work that considers simultaneous exploration and coverage of 3D surfaces of a UAV team \cite{renzaglia2020common}. This is done by constructing a centroidal Voronoi tessellation using MacQueen’s algorithm during the coverage phase and iteratively selecting the closest frontier to navigate to during the exploration phase. However, this approaches address only target search problems without considering simultaneous target tracking.

\subsection{Multi-Target Search and Tracking}
Search and track algorithms must balance the trade-off between exploration (search) and exploitation (tracking), as the information gathering task of the team is prone to getting stuck in local optima due to communication constraints in distributed MRSs.
Multi-robot source-seeking algorithms \cite{du2021multi} aim to maximize information gain by leveraging detected target information, while coverage path planning methods \cite{tang2025large, nair2020mr, senthilkumar2012multi} take the opposite approach, focusing on a pure exploration task with the goal of ensuring that every location on the map is visited by at least one robot in the team.
To overcome the trade-off challenge, some work uses a centralized information-based control strategy that plans feasible trajectories for individual robots then selects the joint set of trajectories that maximizes mutual information between the predicted targets and future detections of the robots \cite{dames2017detecting}.
Kantaros et al.~\cite{kantaros2021sampling} propose a non-myopic distributed sampling-based algorithm that incrementally constructs a directed tree that explores both the information space and the robot motion.
Furthermore, coverage control-based methods \cite{chen2025distributed, ramachandran2023resilient, tolstaya2021multi} integrate collaborative target state estimation with robot trajectory allocation, forming an end-to-end framework from sensor input to control output.
Dames \cite{dames2020distributed} utilizes the estimated target distribution intensity from the distributed PHD filter as weights to update the centroidal Voronoi tessellation using Lloyd's algorithm \cite{cortes2004coverage}. 
This enables a distributed MRS to move toward regions with higher target density while conducting coverage search, naturally balancing exploration and exploitation.

\subsection{Contributions}
This paper adopts the multi-target simultaneous search and tracking framework from \cite{dames2020distributed} and extends it to applications in unknown, non-convex scenarios by incorporating a frontier-based exploration method.
Similar ideas appear in \cite{haumann2013discoverage, bhattacharya2014multi}, which propose that the weight assigned to each point within an exploration region depends on the distance between each robot and the frontier at the edge of its Voronoi cell.
Thus, guiding each robot in its cell using a gradient-based control law allows it to approach the frontier, thereby entering unexplored areas.

In this paper, three primary contributions are made.
First, we develop a novel simultaneous environmental exploration, target search, and target tracking motion planning framework of an unknown and non-convex task space.
This is done by determining whether to use the location with the highest frontier density or with the highest PHD density as the temporary goal for each robot based on whether its dominance region contains the frontier.
Second, the dominant regions of each robot are weighted using normalized unused sensing capacity (NUSC)~\cite{chen2025distributed}, a novel way to optimize spatial allocation and accelerate global convergence.
Third, the capabilities of the proposed strategy in effective target search and tracking are validated through simulation experiments.
To our knowledge, this work achieves for the first time simultaneous full-coverage search and precise tracking of multiple targets with unknown numbers and states using a distributed MRS equipped with limited-range sensors in unknown and non-convex maps.

\section{Problem Formulation}
Consider a network of $n$ mobile robots denoted by $S = \{s_1, \ldots, s_n\}$ with two-dimensional positions $Q = \{q_1, \ldots, q_n\} \subset \mathbb{R}^2$. 
Each robot $s_i$ is equipped with a sensor with finite field of view (FoV) $F_i$. 
We assume that the sensors are isotropic for simplicity, although our algorithm also applies to anisotropic sensors by using the centroid of the detection strategy in \cite{chen2025distributed}.

Robots operate in a partially known, non-convex environment $E \subset \mathbb{R}^2$. 
The environment is made up of two disjoint pieces, $E = E_{\rm known} \cup E_{\rm unknown}$, representing the known and unknown regions of the environment.
The robots aim to explore $E_{\rm unknown}$ while simultaneously searching for and tracking a set of targets in $E_{\rm known}$. The number of targets is unknown and possibly time-varying, and the targets may be static or moving.
The explored environment $E_{\text{known}}$ undergoes continuous updates as robots move and collect new spatial information.

For the sake of simplicity in the discussion, this paper makes the following assumptions.
\begin{assumption}
We assume that robot teams can achieve distributed autonomous localization and mapping using data collected from onboard sensors such as LiDAR and depth cameras, employing frameworks like Door-SLAM \cite{lajoie2020door} and Kimera-Multi \cite{tian2022kimera}. 
Although these methods inevitably introduce localization and mapping errors, this paper assumes these errors are sufficiently small to be negligible to focus the discussion on search and tracking motion control.
\end{assumption}
\begin{assumption}
We assume that the robot can exchange data with all neighbors without packet loss or delay.
In Section~\ref{subsubsec:power diagram}, we define neighbors based on whether the task spaces assigned to two robots are adjacent.
\end{assumption}

During target search and tracking operations, the robot must continuously execute multi-target state estimation algorithms at all times.
In \cite{dames2020distributed}, a multi-robot multi-target state estimation scheme is proposed utilizing the probability hypothesis density (PHD) as the density function $\phi(x)$ in Equation~\ref{eq:lloyd_cost_funtion}.
The PHD, denoted $\nu_t(x)$, represents the spatial density of targets in the search space and is recursively estimated online using a \emph{distributed PHD filter} \cite[Algorithms 1-3]{dames2020distributed}.
The prediction step in the PHD filter accounts for the target motion $f(x|\xi)$, target survival $p_s(\xi)$, and the birth of new targets $b(x)$: 
\begin{equation} 
\label{eq:PHD pre}
\overline{\nu}_t(x) = b(x) + \int_E f(x|\xi) p_s(\xi) \nu_{t-1}(\xi)d\xi.
\end{equation}
The update step accounts for the detection probability $p_d(x|q_i)$ and the measurement errors $g(z|x)$: 
\begin{equation} 
\label{eq:PHD update}
%\nu_t(x) = [1 - p_d(x|q_i)] \overline{\nu}_t(x) + \sum_{z_i \in Z_t} \frac{\psi_{z_i,q_i}(x) \overline{\nu}_t(x)}{\eta_{z_i}(\overline{\nu}_t)},
\nu_t(x) = [1 - p_d(x|q_i)] \overline{\nu}_t(x) + \sum_{z \in Z_t} \frac{p_d(x|q_i) g(z|x) \overline{\nu}_t(x)}{\eta_{z}(\overline{\nu}_t)},
\end{equation} 
where $g(z|x)$ represents the measurement likelihood function that defines the probability density of observing measurement $z$ given a target exists at position $x$, $Z_t$ is the set of all measurements at time $t$ and $\eta_{z}(\overline{\nu}_t)$ is the normalization term in the PHD filter. 
This recursive approach enables effective multi-target tracking in dynamic environments without requiring explicit data association.
Utilizing the output of the PHD filter as the importance weighting function within the standard Lloyd’s algorithm, i.e., setting $\phi(x) = \nu_t(x)$, causes the robots to be drawn toward areas that are likely to contain targets. 
This yields effective search and tracking behavior of an unknown and time-varying number of static or moving targets.

\section{Simultaneous Search and Tracking}

\begin{figure}[tbp]
    \centering
    \includegraphics[width=1.0\linewidth]{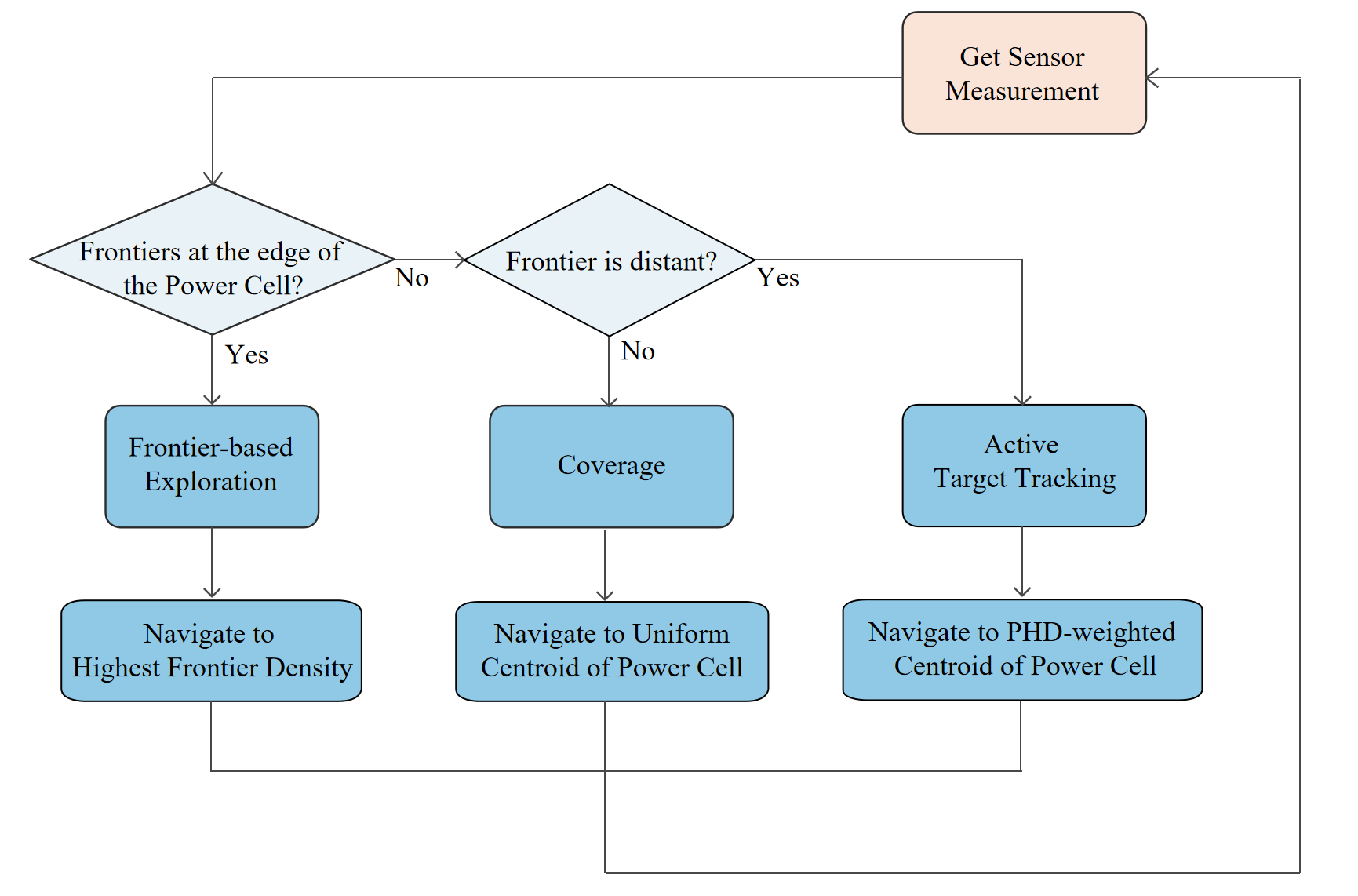}
    \caption{Figure shows the overall algorithmic flow.}
    \label{fig:System_Diagram}
\end{figure}

The general framework of our system, shown in Figure \ref{fig:System_Diagram}, contains three control strategies: frontier-based exploration, coverage for target search, and active target tracking. 
The selection of control strategies is based on the presence or absence of the frontier and its relative distance from the robot.
Through such a mechanism, the multiple task objectives of exploration, search, and target tracking can be optimized simultaneously.
We shall elaborate on these three strategies and the overall algorithmic framework in detail in subsequent sub-sections.
During the execution of these three strategies, the robots utilize a distributed PHD filter at all time steps for multi-target state estimation.

\subsection{Optimized Space Assignment using Power Diagram}
Before introducing the robot motion control strategy, we first outline the robot spatial allocation method, which serves as the prerequisite for our collaborative control algorithm framework.
Our work builds on the strategy for multi-target simultaneous search and tracking in known convex environments by utilizing a Lloyd's algorithm-based framework.
We adopt a variant of Lloyd's Algorithm \cite{cortes2004coverage} that aims at minimizing the locational cost function 
\begin{equation} 
\label{eq:lloyd_cost_funtion}
H(\mathcal{Q}, \mathcal{W}) = \sum_{i=1}^{n} \int_{\mathcal{W}_i} \big(\|x - q_i\|^2 - r_i^2\big) \phi(x)~dx, 
\end{equation} 
with respect to both the locations of the generating points (i.e., robots) $\mathcal{Q}$ and the assigned dominant regions $\mathcal{W}$ of them, where $\mathcal{W}_i$ is the space assigned to robot $s_i$, $r_i$ is the additive weight (or power radius) of $s_i$, $\|\cdot\|$ is Euclidean norm and $\phi(x)$ is a density function over the task space.

\subsubsection{Power Diagram}
\label{subsubsec:power diagram}
Minimizing $H$ with respect to $\mathcal{W}$ yields the power cell $P_i$ of each robot $s_i$, given by
%%%%%%%%%%%%%%%%%%%%%%%%%%%%%%%%%
\begin{equation}
\label{eq:power_diagram_definition}
P_i = \{x \in E_{\text{known}}~|~\|x - q_i\|^2 - r_i^2 \leq \|x - q_j\|^2 - r_j^2\},
\end{equation}
%%%%%%%%%%%%%%%%%%%%%%%%%%%%%%%%%
where $s_j$ is any robot other than $s_i$ in $S$. 
The power radii $r_i$ are the key parameters of this method.
Based on the definition of $P_i$, we define \emph{neighbors} of $s_i$ as $\mathcal{N}_i = \{s_j~|~\forall s_j\in S, P_j~\textrm{shares an adjacent edge with}~P_i \}$.

\subsubsection{Normalized Unused Sensing Capacity}
Throughout the time, the team aims to dynamically allocate the explored region $E_{\text{known}}$ optimally among robots based on their current remaining observational capabilities, thus balancing the sensing workload across the robot team.
Therefore, we set $r_i$ as the normalized unused sensing capacity (NUSC), proposed in \cite{chen2025distributed}, which computes the difference between a sensor's theoretical maximum sensing capacity and its current usage, quantifying the current residual information-gathering potential of it.
The maximum sensing capacity of the robot $s_i$' is defined as
\begin{equation}
    C_{\textrm{max},i} = \mu \max_{x \in F_i} p_d(x|q_i) \nu_t(x),
\end{equation}
where $p_d(x|q_i)$ is the detection probability of $s_i$ at location $q_i$, and $\mu$ is a tuning parameter associated with the maximum target density in the task space. 
The expected number of target detections is given by
\begin{equation}
C_{\textrm{exp},i} = \frac{\int_{F_i} p_d(x|q_i) \nu_t(x) \,dx}{\int_{F_i} \nu_t(x) \,dx}.
\label{eq:det_exp}
\end{equation}
Note that $\int_{F_i}\nu_t(x)\,dx$ is numerically equal to the expected number of targets in $F_i$.

Thus, the relative NUSC with respect to the maximum target density for the robot $s_i$, denoted $U_i$, is given by
\begin{equation}
U_i = C_{\textrm{max},i} - C_{\textrm{exp},i}.
\label{eq:unused capacity}
\end{equation}
$U_i$ captures the intuitive notion that sensors operating in regions with high unexplained target density have greater potential for information gathering, while sensors in thoroughly explored or low-density regions exhibit lower unused capacity.
Positive NUSC values indicate regions where continued sensing operations are likely to yield valuable information, while values approaching zero suggest areas where sensing resources might be more effectively deployed elsewhere.

We then use the NUSC to set the power radii $r_i$ so that the power diagram accounts for the heterogeneity of the instantaneous workload and the sensing capacity. 
The intuition is that when a robot observes a number of targets that exceed its observational capacity, it indicates a reduced ability to cover and search for other targets. 
Consequently, the assigned coverage area should be reduced to encourage other robots to move closer to its vicinity.

\subsubsection{Power Cell Optimization}
We first assume that $E_{\text{known}}$ is convex.
For robot $s_i$ positioned at $q_i$ with NUSC $U_i$, the optimized space assigned to it is the power cell $P_i$ given by
%%%%%%%%%%%%%%%%%%%%%%%%%%%%%%%%%
\begin{equation}
\label{eq:power_diagram_nusc}
P_i = \{x \in E_{\text{known}}~|~\|x - q_i\|^2 - U_i^2 \leq \|x - q_j\|^2 - U_j^2\}.
\end{equation}
%%%%%%%%%%%%%%%%%%%%%%%%%%%%%%%%%
The advantage of this spatial allocation approach is that it ensures comprehensive target search coverage of the explored areas while aligning the distribution of search formations with the distribution of detected targets. 
This enables faster discovery of undetected targets and reduces the probability that robots will dominate spaces beyond their own exploration capabilities, thereby preventing other robots from conducting effective exploration coverage around them.
Our quantitative results confirm that this will accelerate the speed of global target search.

\begin{figure}[tbp]
    \centering
    \includegraphics[width=0.6\linewidth]{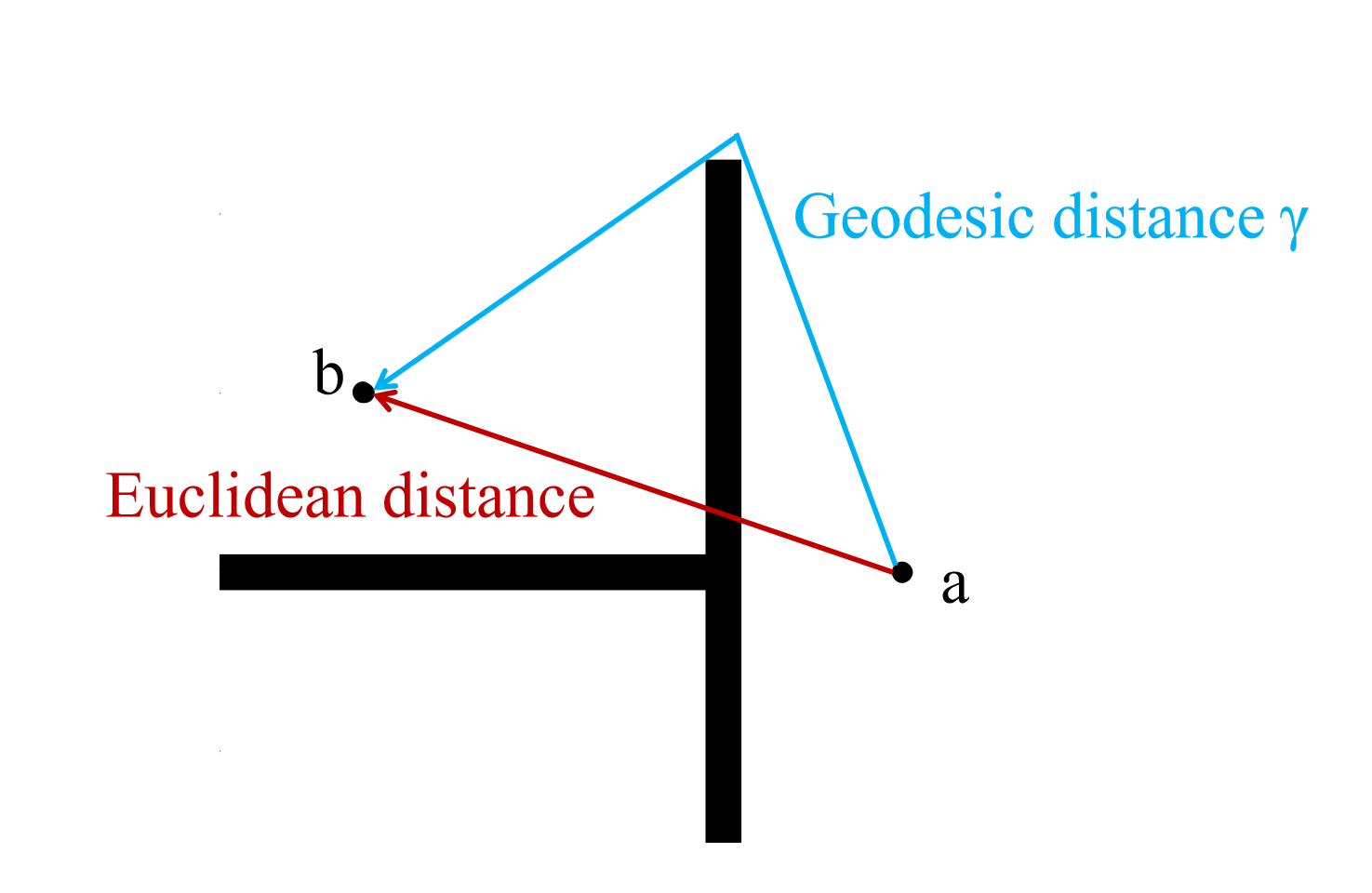}
    \caption{Figure illustrates the geodesic distance. The black dots $a$ and $b$ represent the starting and ending points of a pair of robotic movements, respectively. Black solid lines denote obstacles. Colored lines plot Euclidean distance and geodesic distance, respectively.}
    \label{fig:geodesic_distance}
\end{figure}

To adapt to non-convex environments, $\|\cdot\|$ in Equation~\ref{eq:power_diagram_nusc} must be replaced with the geodesic distance, which measures the shortest path \textit{fully contained inside a space} between two points \cite{bhattacharya2014multi}.
For a path $\gamma(t), \, t \in [0,1]$, this is defined as $d_E(a,b) = \inf\{L(\gamma)~|~\gamma(t) \in E_{\rm known} \, \forall t \in [0,1], \; \gamma(0) = a, \; \gamma(1) = b\}$, shown in Figure \ref{fig:geodesic_distance}.
In practical applications, the task space can be gridded and search-based path planning algorithms such as Dijkstra's algorithm and Floyd-Warshall algorithm \cite{hougardy2010floyd} can be used to generate the shortest path between the robot's grid cell and other grid cells, thereby calculating the geodesic distance.
%%%%%%%%%%%%%%%%%%%%%%%%%%%%%%%%%
%%%%%%%%%%%%%%%%%%%%%%%%%%%%%%%%%

\subsection{Optimizing Locations}
Once optimal space allocation is achieved, we need to develop control strategies to drive robots toward the most ideal locations, enabling rapid environment exploration, optimal coverage search, and target following for accurate tracking.
The following sections discuss three locational optimization strategies by determining a temporary goal to drive to within the power cell for each robot.
These control strategies are mutually coupled and synergistically optimized. 
By employing the appropriate control strategy at the suitable stage, the robot can simultaneously achieve efficient environmental exploration and target search and tracking.

\subsubsection{Strategy 1: Frontier-based Multi-robot Exploration}
\label{sec:multi-robot-exploration}
Unlike traditional multi-robot exploration problems, in this paper, multi-robot exploration aims not only to rapidly map uncharted space but also to collaboratively search for and estimate the states of multiple targets.
To ensure that more targets are covered during the exploration phase, our cooperative exploration strategy integrates \emph{optimized space assignment} with \emph{locational optimization} utilizing a Lloyd's algorithm framework, enabling robots to simultaneously perform environmental exploration and coverage for target tracking.

%In robotic mapping problems, a frontier of ${E}_{\text{known}}\in E$ is define by $L= \overline{E}_{\text{known}} \setminus {E}_{\text{known}}^\circ$, where $\overline{E}_{\text{known}}$ is the closure of ${E}_{\text{known}}$ and ${E}_{\text{known}}^\circ$ is the interior of ${E}_{\text{known}}$.
In robotic mapping problems, a frontier of ${E}_{\text{known}}\in E$ is a set of points between ${E}_{\text{known}}$ and ${E}_{\text{unknown}}$, defined by $L = \partial E_{\text{known}} \cap E_{\text{free}}$, where $\partial E_{\text{known}}$ represents the boundary of $E_{\text{known}}$ and $E_{\text{free}}$ is the obstacle-free space within $E_{\text{known}}$.
We denote by $l_i$ the set of frontier points on the edge of $P_i$ such that $L=l_1 \cup l_2 \cup \ldots \cup l_n$.
These frontiers are critical for efficient environmental discovery as they represent areas where new information can be acquired.

When a non-empty set of frontiers $l_i$ is found on the boundary of the power cell $P_i$ of a robot $s_i$, it adopts a frontier-based exploration control strategy to prioritize mapping unknown regions. 
In \cite{renzaglia2020common}, a robot is iteratively driven towards the closest frontier in $L$.
However, this does not maximize the exploration gains obtained by the robot with each step it takes.
To overcome this limitation, in this paper, our proposed exploration strategy iteratively drives each robot towards the location where the density of frontier points at the edge of its power cell is maximal, i.e., the temporary goal of the robot $s_i$ at each time step is given by
%%%%%%%%%%%%%%%%%%%%%%%%%%%%%%%%%
\begin{equation}
\label{eq:frontier_goal_detailed}
g_{\text{exp},i} = \argmax_{q_i \in P_i} \rho_{\text{ftr}}(q_i),
\end{equation}
%%%%%%%%%%%%%%%%%%%%%%%%%%%%%%%%%
where $\rho_{\text{ftr}}(q_i)$ is the frontier density obtained by using kernel density estimation (KDE) with a Gaussian kernel, i.e.,
%%%%%%%%%%%%%%%%%%%%%%%%%%%%%%%%%
\label{eq:frontier_density_detailed}
\begin{equation}
\rho_{\text{ftr}}(q_i) = \frac{1}{|l_i|} \sum_{j=1}^{|l_i|} \frac{\exp\left( -\frac{(q_{i,x} - l_{j,x})^2}{2 h_x^2} - \frac{(q_{i,y} - l_{j,y})^2}{2 h_y^2}\right)}{2 \pi h_x h_y},
\end{equation}
%%%%%%%%%%%%%%%%%%%%%%%%%%%%%%%%%
where $q_i = (q_{i,x}, q_{i,y})$ is the robot location $s_i$, $l_i$ is the set of frontier points at the edge of the $s_i$'s power cell $P_i$, $|l_i|$ is the number of frontier points, $l_{j,x}$ and $l_{j,y}$ are the x- and y-coordinates of the $j$-th frontier point, respectively, and $h_x$, $h_y$ are the bandwidth parameters for the x-axis and y-axis, respectively, which control the smoothness of the kernel function.
The density is obtained by summing Gaussian kernel contributions from all grid points on frontiers, with each point's influence decaying with distance.
%The KDE with a Gaussian kernel provides a flexible, non-parametric method for estimating densities in continuous spaces. 
Therefore, higher weights are assigned to larger frontier clusters, ensuring that extensive unexplored regions receive prioritized attention over isolated frontier points, maximizing exploration efficiency per robot movement. 
%The bandwidth parameters $h_x$ and $h_y$ are determined using the Silverman rule of thumb, providing adaptive kernel width selection based on the density of the frontier.
%It adapts well to unknown environments, allowing robots to efficiently explore frontiers without needing prior knowledge of the structure of the space.

This approach maximizes the exploration gain per step of the robot's movement, ensuring each exploring step covers the maximum number of frontiers from the previous step. 
However, as the location of the highest-density frontier fluctuates, this also causes the robot's temporary goal to sometimes exhibit significant jumps, resulting in pronounced turning or backtracking movements. 
Although this improves coverage search effectiveness, it may also slow down exploration speed. 
We will discuss this in the qualitative results section.

\subsubsection{Strategy 2: Cooperative Coverage for Target Search}
\label{sec:coverage}
Minimizing $H$ in Equation~\ref{eq:lloyd_cost_funtion} with respect to $\mathcal{Q}$ drives each robot $s_i$ to the centroid of $P_i$ weighted by the density function $\phi(x)$.
The choice of $\phi(x)$ determines the motion behavior of the team, e.g., spreading out for area coverage, or active target following for accurate tracking.
In the cases where no frontiers are located at the edge of the power cell $P_i$ of the robot $s_i$ in the exploration phase, we set $\phi(x) = 1, \forall x\in E_{\text{known}}$.
The uniform density function allows each robot to perform a control strategy of Lloyd's-based coverage for searching targets within the explored regions. 
The robot is iteratively driven towards the temporary goal given by
%%%%%%%%%%%%%%%%%%%%%%%%%%%%%%%%%
\begin{equation}
\label{eq:coverage goal}
g_{\text{cov},i} = \argmin_{q_i \in P_i} \max_{x \in P_i} d_E(q_i, x),
\end{equation}
%%%%%%%%%%%%%%%%%%%%%%%%%%%%%%%%%
which represents the geometric centroid of $P_i$, which is the point that minimizes the maximum distance to any point within the power cell.
The coverage strategy ensures that no point within the assigned territory is excessively far from the robot's position, providing uniform coverage quality across the entire power cell.

This strategy maintains optimal spatial coverage while keeping robots responsive to exploration opportunities that may emerge in neighboring regions. 
By implementing a Lloyd's algorithm-based framework, the strategy optimizes coverage geometry within assigned territories, ensuring that robots maintain strategic positioning for future exploration phases while recursively updating PHD within their power cells for target state estimation.

%Strategies 1 and 2 allow the team to rapidly map unexplored space while collaboratively searching for and estimating multiple target states based on the time-varying heterogeneous sensing capability of each individual. 
%Innovation lies in combining frontier-based exploration algorithms with coverage control based on Lloyd's algorithm.  
%Each exploration step maximizes information gain during exploration, while coverage efficiency is enhanced through optimized task allocation in coverage search.

\subsubsection{Strategy 3: Active Target Tracking}
\label{sec:simultaneous_tracking}
As the explored area continues to expand, robots performing coverage search tasks within the explored region become increasingly sparse, making it difficult to maintain enough density to effectively cover all areas. 
Consequently, robots must adjust their spatial distribution by following regions with high target density to optimize overall target tracking accuracy.
At this point, robots enter the active multi-target tracking phase, employing a control strategy that involves cooperatively moving to track the targets more accurately.

Strategy 3 builds upon the distributed PHD filtering framework established in \cite{dames2020distributed}, adapting it to non-convex environments through replacing the Euclidean distance in constructing a power diagram with the geodesic distance.
During this phase, robots execute sensor-based motion planning that prioritizes regions with higher target presence probability, following the temporary goal given by
%%%%%%%%%%%%%%%%%%%%%%%%%%%%%%%%%
\begin{equation}
\label{eq:tracking goal}
%g_{\text{track},i} = \frac{\sum_{j: s_j \in P_i} \nu_t(x) \cdot s_j}{\sum_{j: s_j \in P_i} \nu_t(x)}
g_{\text{track},i} = \frac{\int_{P_i} \nu_t(x) \cdot x \, dx}{\int_{P_i} \nu_t(x) \, dx},
\end{equation}
%%%%%%%%%%%%%%%%%%%%%%%%%%%%%%%%%
where $\nu_t(x)$ represents the PHD function indicating target presence probability density within robot $s_i$'s spatial assignment.
This approach follows the distributed PHD filtering paradigm, where the estimated target distribution intensity serves as weights for trajectory planning, enabling precise target localization while maintaining distributed coordination properties.
Note that since the space is non-convex, the weighted centroid of a power cell may fall outside the cell. 
To ensure the safety of the robot's motion planning, we replace any temporary goal that falls outside the cell with the point in that cell closest to it.

\subsubsection{Transitions Between Strategies}
As illustrated in Figure~\ref{fig:System_Diagram}, robots transition between strategies depending on whether there is a frontier nearby.
When there are frontier points at the boundary of the power cell, the robot should take Strategy 1 to prioritize environmental exploration to draw the complete map as quickly as possible, thus ensuring the eventual optimization of coverage and tracking.
If there is no frontier at the edge of the cell, but there are frontiers nearby that are closer to it, the robot should adopt Strategy 2 to prioritize coverage. 
This prevents the delay in finding undetected targets in the newly explored area due to following the target within previously explored regions, while also avoiding hindering the path of exploring robots.

We still need to determine under what conditions it would be appropriate to trigger Strategy 3.
Intuitively, when the density of robots in the explored area becomes sufficiently low-to the point where it neither provides adequate coverage of the area nor hinders exploration robots from moving toward the frontiers—robots should employ Strategy 3 to move alongside the targets.
This can be converted into determining whether a frontier exists within a certain range $d_f$ around a robot, where $d_f$ is a tuning parameter related to the sensor's coverage range.
If $d_f$ is too small, the robots begin tracking the targets when concentrated in a small area, prematurely converging on local maxima and compromising the effectiveness of the coverage search. 
Conversely, if $d_f$ is too large, robots do not adopt target following strategies even when the density is too low for effective coverage search, resulting in loss of tracking accuracy.
Setting the distance to approximately twice the sensor's observation diameter is a suitable choice, as this allows the field of view of a robot using Strategy 2 to cover the entire search area between robots using Strategy 1 to explore the frontier and robots using Strategy 3 to follow targets.

\subsection{Algorithm Outline}

%%%%%%%%%%%%%%%%%%%%%%%%%%%%%%%%%
\begin{algorithm}[tbp]
\caption{Distributed Search and Tracking}
\label{alg:comprehensive_framework}
\begin{algorithmic}[1]
\small
\STATE Initialization step executed
\WHILE{mission active}
    \FOR{each robot $s_i \in \mathcal{S}$ \textbf{in parallel}}
        \STATE Get sensor measurement at $q_i$    
        \STATE Send $q_i$ to and receive $q_j$ from all neighboring $s_j$
        \STATE Update $\nu(x)$, $U_i$, $P_i$, $l_i$    
        \IF{$l_i \neq \emptyset$}
            \STATE $\beta_i \leftarrow \min_{x \in l_i} \|q_i - x\|$
            \STATE $ g_{temp,i} \leftarrow g_{\text{exp},i}$ {via Equation (\ref{eq:frontier_goal_detailed}})
            \STATE Send $\beta_i$ to all neighboring robots $s_j \in \mathcal{S}$   
        \ELSE
            \STATE Receive $\beta_j$ from all neighboring robots $s_j \in \mathcal{S}$
            \STATE $\beta_i \leftarrow \min_{x \in \beta_j} \|q_i - x\| $ for all neighboring $s_j$
            \STATE Send $\beta_i$ to all neighboring robots $s_j \in \mathcal{S}$   
            \IF{$\beta_i > d_f$}  
                \STATE $g_{temp,i} \leftarrow g_{\text{track},i}$ via  Equation (\ref{eq:tracking goal})
            \ELSE
                \STATE $g_{temp,i} \leftarrow g_{\text{cov},i}$ via Equation (\ref{eq:coverage goal})
            \ENDIF
        \ENDIF
        \STATE $g_i \leftarrow g_{temp,i}$ 
    \ENDFOR    
    \STATE $t \leftarrow t + 1$
\ENDWHILE
\end{algorithmic}
\end{algorithm}
%%%%%%%%%%%%%%%%%%%%%%%%%%%%%%%%%

Our multi-robot simultaneous multi-target search and tracking algorithm for an unknown and non-convex environment is outlined in Algorithm~\ref{alg:comprehensive_framework}. 
The team must first initialize the space assignment by constructing a power diagram weighted by $C_{max,i}$ of each robot $s_i$ over the region covered by their field of view.
The PHD is initialized as uniform, meaning that no prior knowledge of the target distribution is known.
The initialization step follows Algorithm 1 demonstrated in \cite{chen2025distributed}.
Then, at each iteration, every robot $s_i$ operates in parallel by acquiring sensor measurements and exchanging position information $q_i$ with neighboring robots for distributed power cell update, as outlined in Line 5. 
Each robot then updates its local PHD, NUSC weights $U_i$, power cell $P_i$, and local set of frontiers $l_i$.

The core component of the motion planning algorithm is a three-strategy decision mechanism based on whether the frontiers are located at the edge of the power cell and the robot's minimum distance to the frontier.
When local frontiers exist at the edge of the power cell ($l_i \neq \emptyset$), robots execute Strategy 1 in Lines 7-9, computing its minimum distance to a frontier point, denoted by $\beta_i$, and navigating toward maximum frontier density regions using KDE with Gaussian kernel.
When no local frontiers are present, each robot receives frontier distances $\beta_j$ from all neighboring robots as outlined in Line 12 and updates its minimum frontier distance $\beta_i$ in Line 13.
If $\beta_i$ exceeds $d_f$, Strategy 3 is activated, directing robots toward PHD-weighted centroids for target following, with fallback to the coverage strategy when the calculated goals fall outside of the power cells.
Otherwise, Strategy 2 executes, maintaining optimal coverage through spatial centroid navigation, as outlined in Lines 15-19.
The algorithm concludes each iteration with frontier distance broadcasting to neighboring robots as outlined in Lines 10 and 14 in case any neighboring robots request it to compute the minimum distance to the frontier.
Note that this algorithm represents the main iterative loop executed after system initialization (i.e., for $t > 1$), with time indices omitted for clarity, since each iteration represents a single discrete time step.

The distributed algorithm exhibits favorable scalability characteristics with computational complexity comprising power diagram construction $O(\log |N_i|)$ per robot, where $|N_i|$ is the number of neighbors of each robot $s_i$. 
Communication requirements involve only position sharing and frontier distance exchange between neighboring robots. 
Both complexity components scale with local neighborhood size rather than global parameters, ensuring constant per-robot requirements independent of environment size or robot count, making the framework suitable for large-scale deployments in resource-constrained systems.

A graphical representation of Algorithm~\ref{alg:comprehensive_framework} is referred to Figure~\ref{fig:System_Diagram}. 
This comprehensive framework enables seamless transitions between exploration, coverage, and tracking behaviors while maintaining distributed coordination and ensuring exploration completeness through frontier-based coverage, collision-free motion planning through geodesic distance-constrained power diagrams, and cooperative target tracking through distributed PHD filtering.
It simultaneously optimizes three objectives: fast area exploration, comprehensive target search, and accurate target tracking.

%%%%%%%%%%%%%%%%%%%%%%%%%%%%%%%%%%%%%%%%%%%%%%%%%%%%%%%%%%
\section{Results}
%%%%%%%%%%%%%%%%%%%%%%%%%%%%%%
\begin{figure*}[tbp]
    \centering
    \subfloat[500s]{
    	\includegraphics[width=0.31\columnwidth]{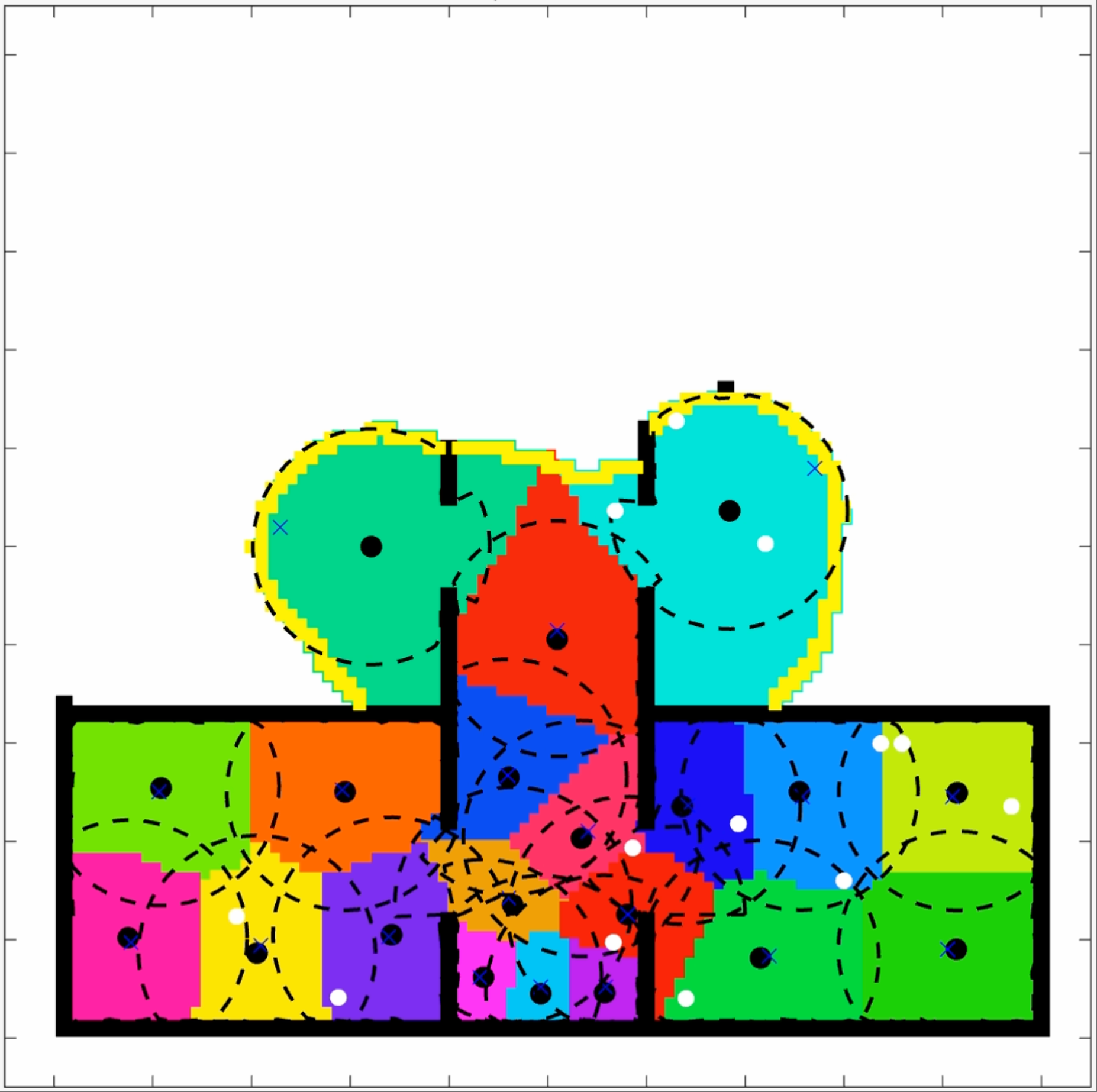}\label{fig:500p}
    }
    \subfloat[1000s]{
    	\includegraphics[width=0.31\columnwidth]{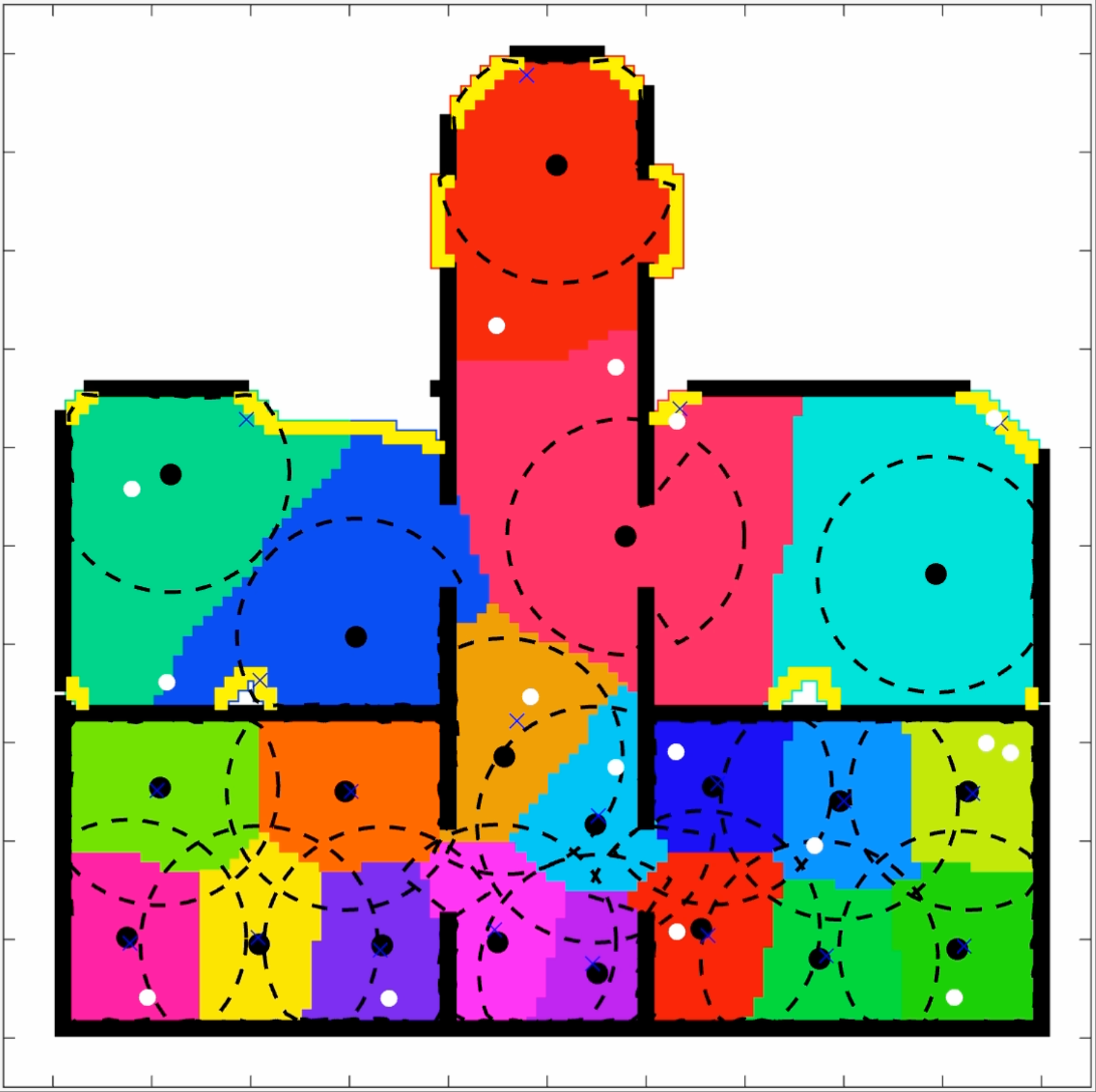}\label{fig:1000p}
    } 
    \subfloat[1500s]{
    	\includegraphics[width=0.31\columnwidth]{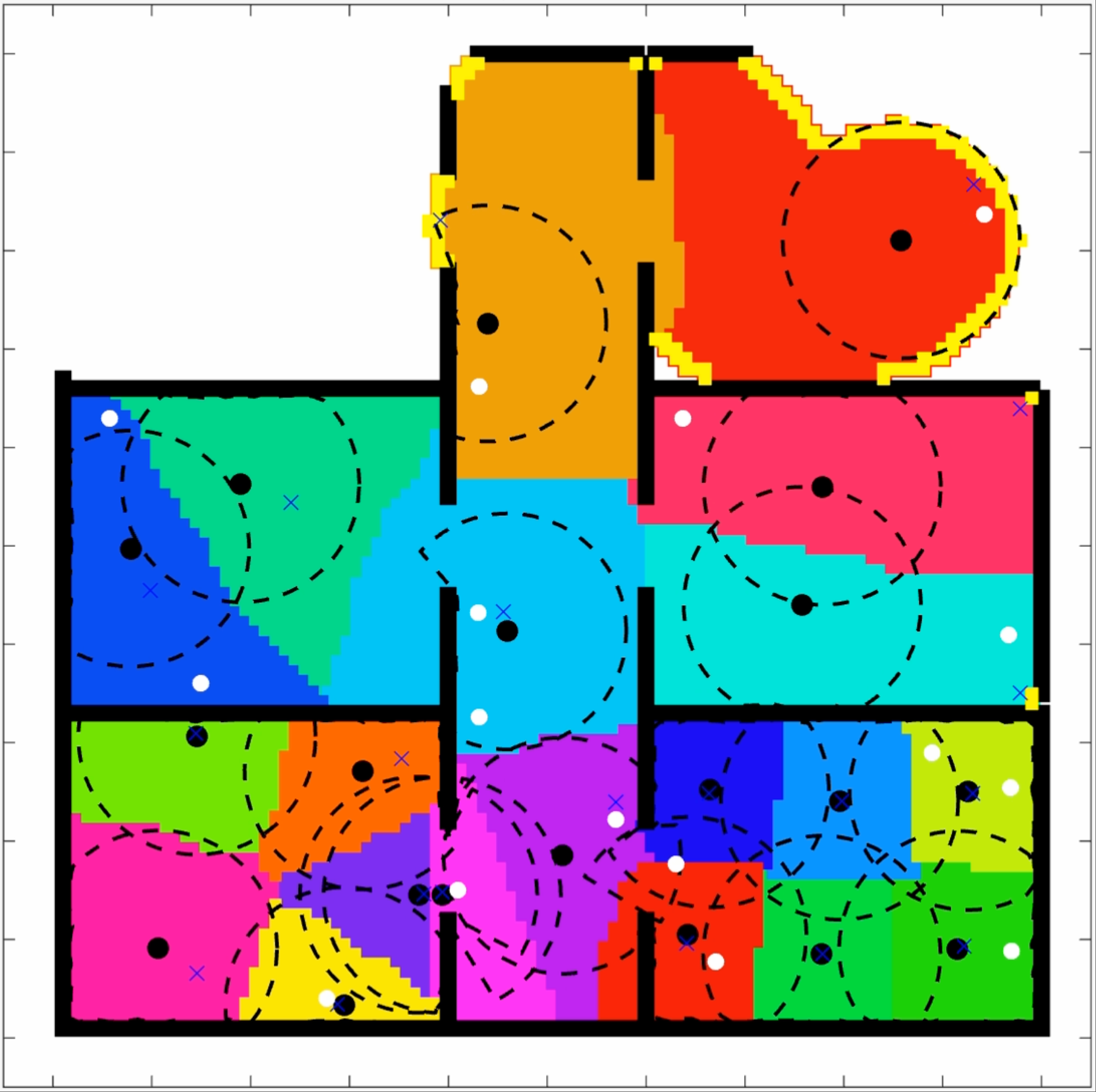}\label{fig:1500p}
    } 
    \subfloat[2000s]{
    	\includegraphics[width=0.31\columnwidth]{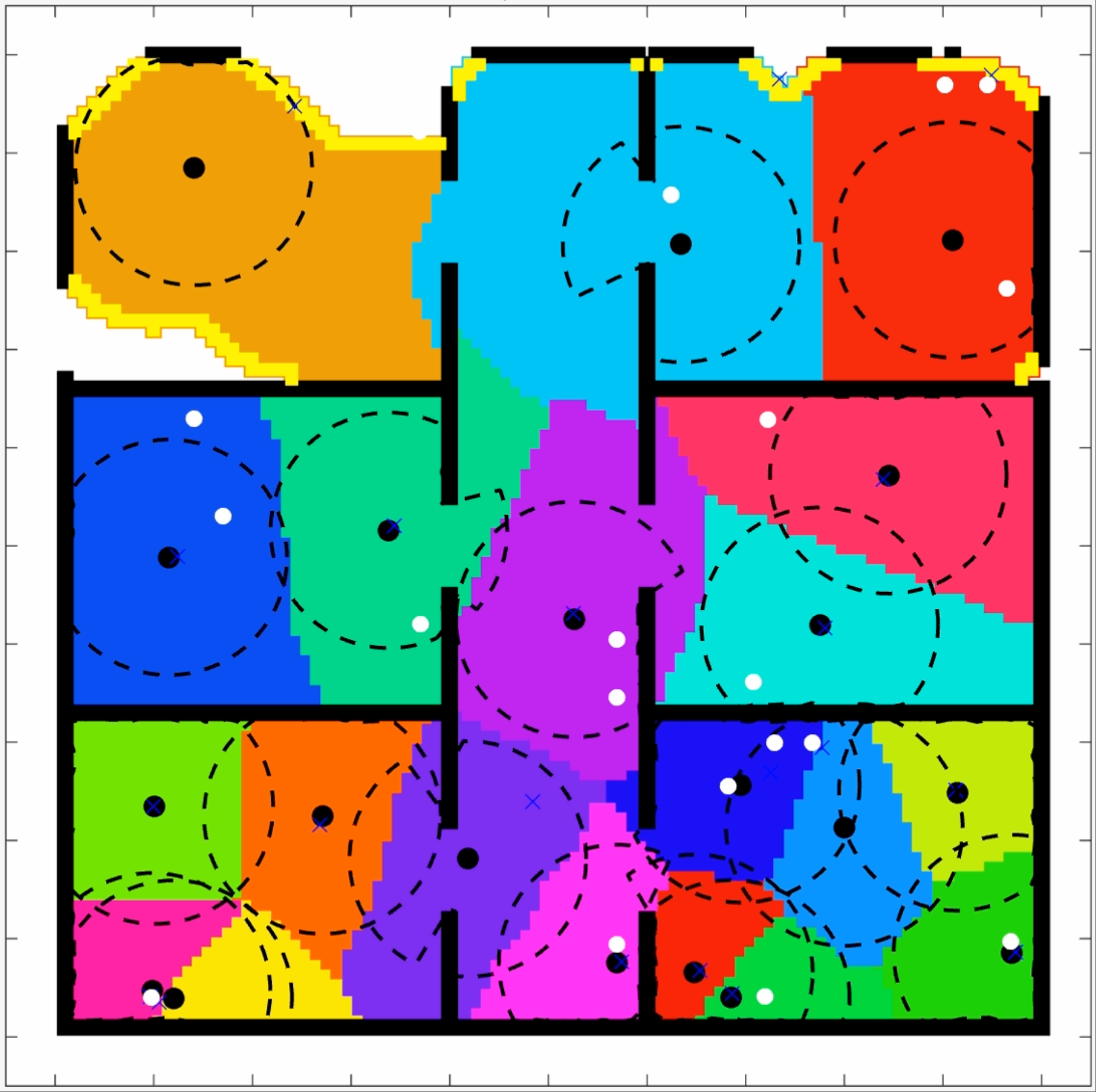}\label{fig:2000p}
    } 
    \subfloat[3000s]{
    	\includegraphics[width=0.31\columnwidth]{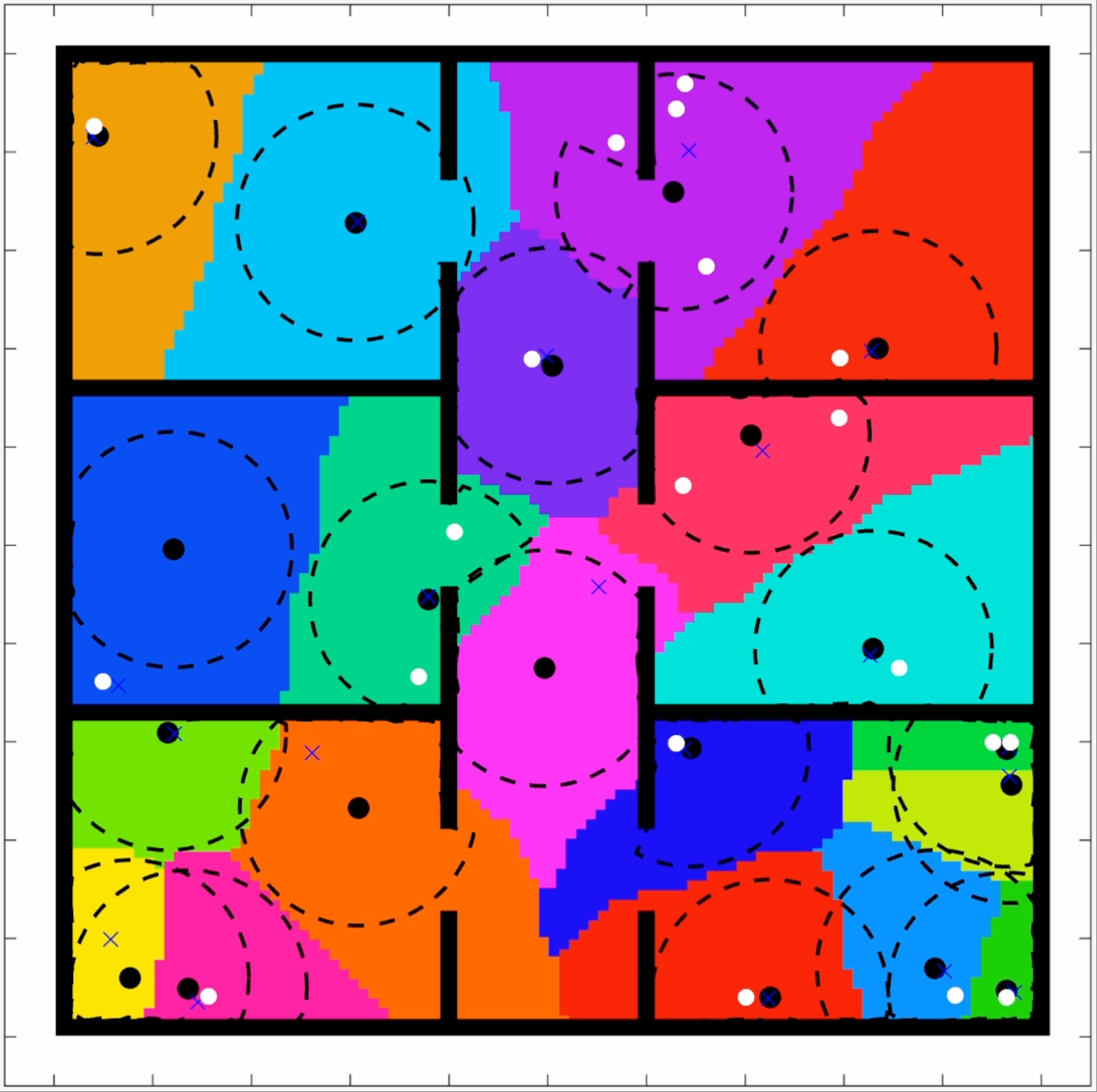}\label{fig:3000p}
    }
    \subfloat[4000s]{
    	\includegraphics[width=0.31\columnwidth]{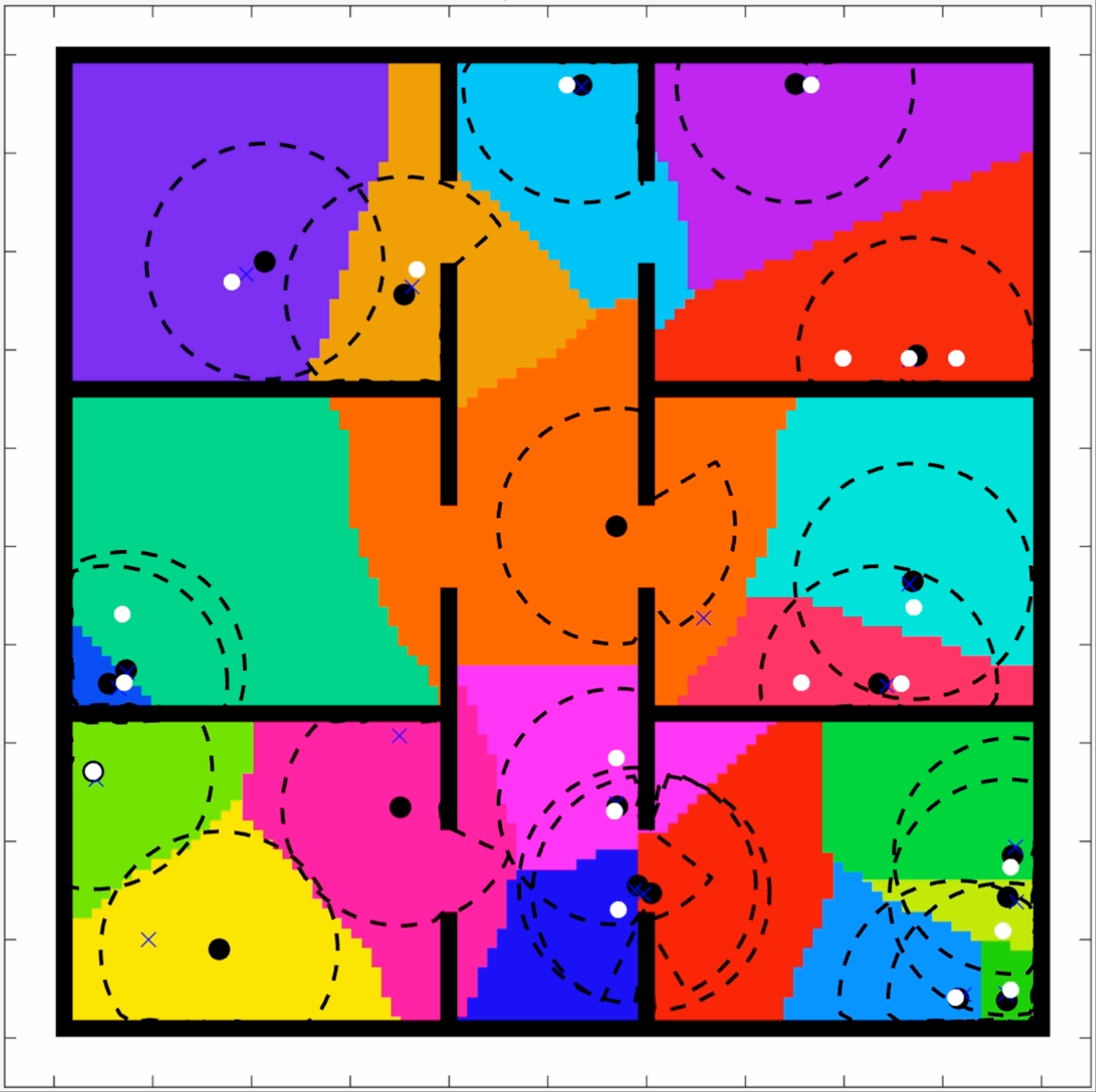}\label{fig:4000p}
    } \\
    \subfloat[500s]{
        \includegraphics[width=0.31\columnwidth]{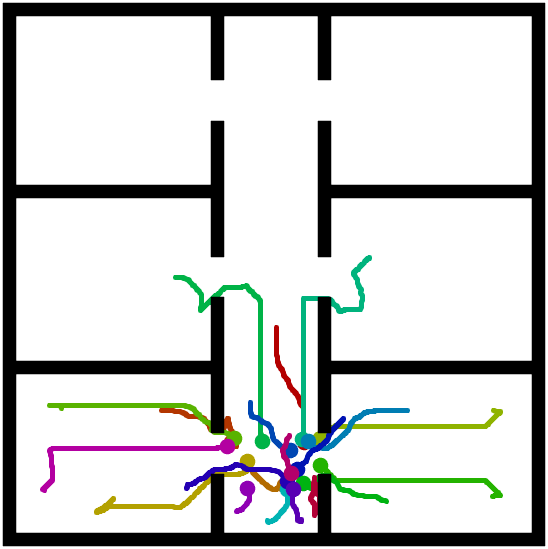}\label{fig:500}
    } 
    \subfloat[1000s]{
    	\includegraphics[width=0.31\columnwidth]{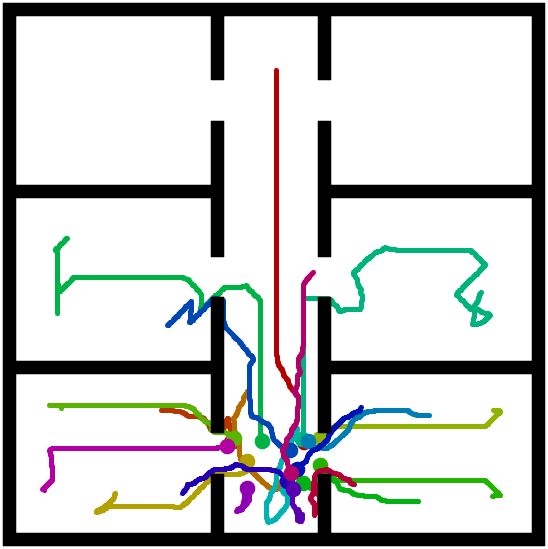}\label{fig:1000}
    } 
    \subfloat[1500s]{
    	\includegraphics[width=0.31\columnwidth]{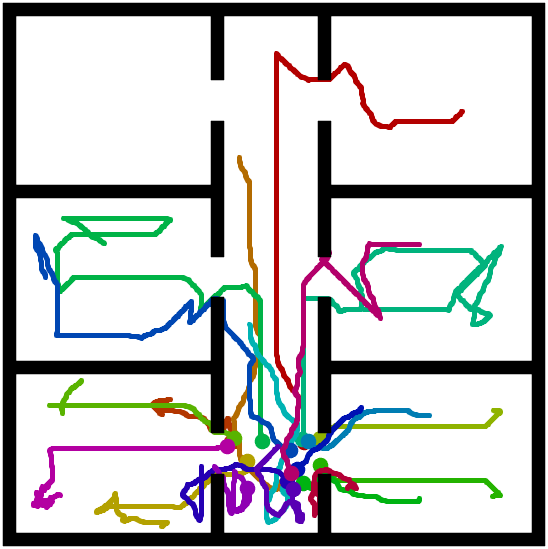}\label{fig:1500}
    } 
    \subfloat[2000s]{
        \includegraphics[width=0.31\columnwidth]{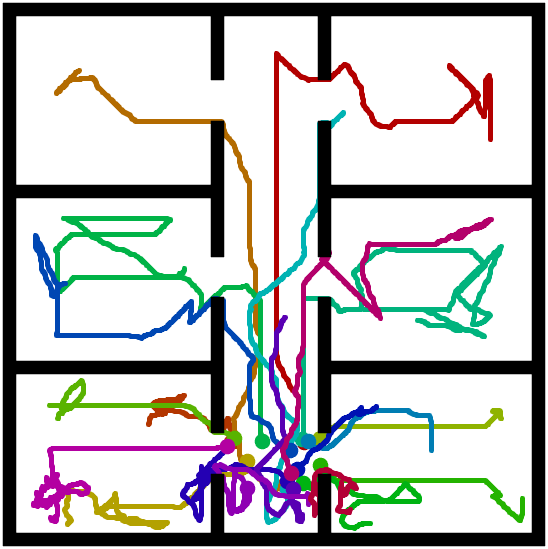}\label{fig:2000}
    } 
    \subfloat[3000s]{
    	\includegraphics[width=0.31\columnwidth]{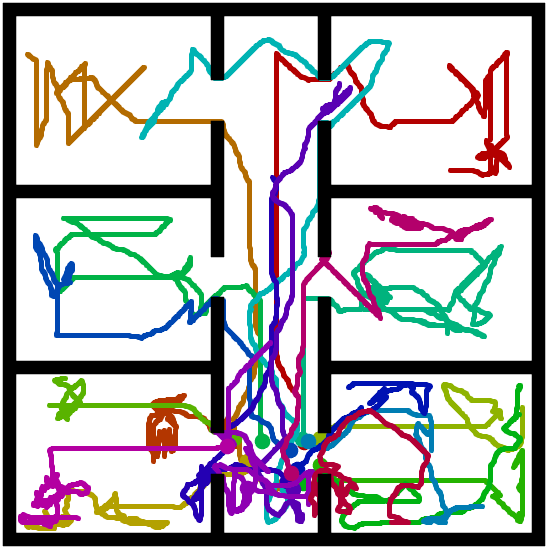}\label{fig:3000}
    } 
    \subfloat[4000s]{
    	\includegraphics[width=0.31\columnwidth]{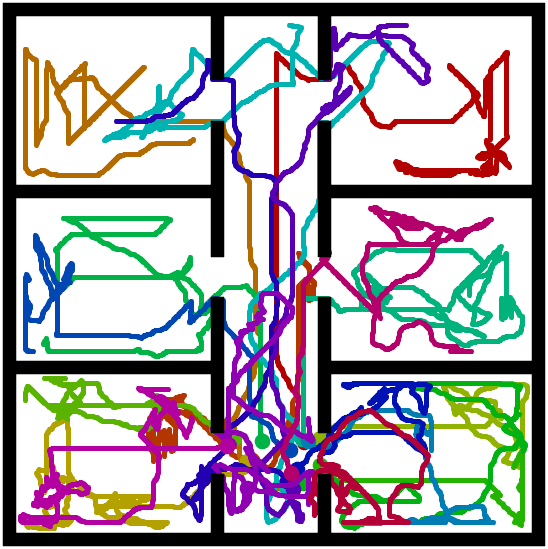}\label{fig:4000}
    }
    \caption{The set of figures depicts a scenario where 20 robots search for and track 20 moving targets within a 100m x 100m indoor map, showing six time points during a 4000-second mission. In the first two rows, thick black solid lines represent explored map areas. White dots indicate target locations. Black dots and dashed circles denote robot positions and fields of view. Yellow borders indicate frontiers. Crosses mark robots' temporary goals. Colored regions represent each robot's power cell.
    The two rows below plot the paths traversed by each robot at different times on the map, distinguished by different colors. The colored circle at the bottom center of the map represents the starting position of the corresponding robot.}
    \label{fig:traj}
\end{figure*}
%%%%%%%%%%%%%%%%%%%%%%%%%%%%%%

%%%%%%%%%%%%%%%%%%%%%%%%%%%%%%
\begin{figure}[tbp]
\centering
\subfloat[Partition Comparison]{
\includegraphics[width=0.64\columnwidth]{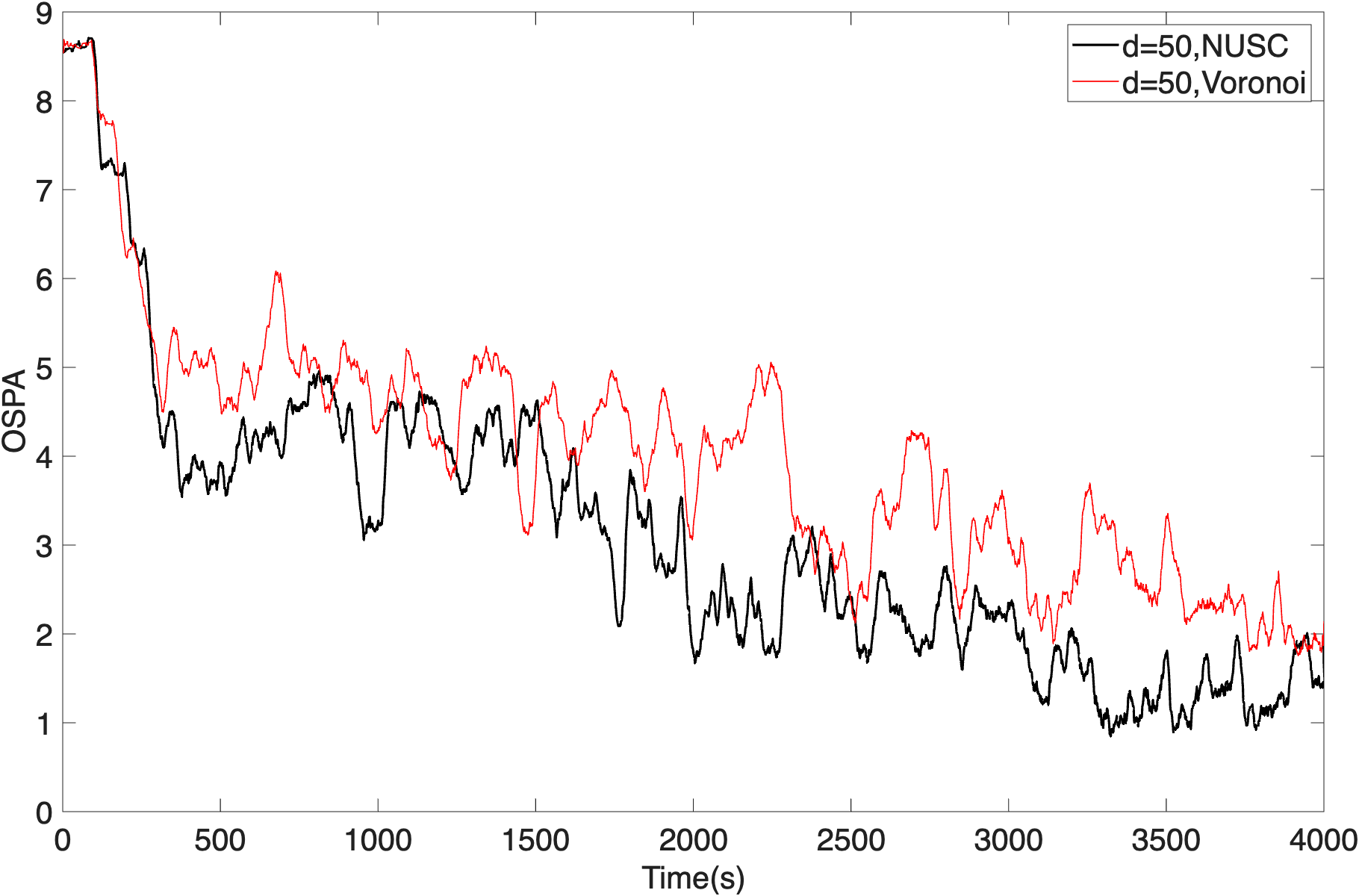}
\label{fig:voronoi_comparison}
} \\
\subfloat[$d_f$ Comparison]{
\includegraphics[width=0.64\columnwidth]{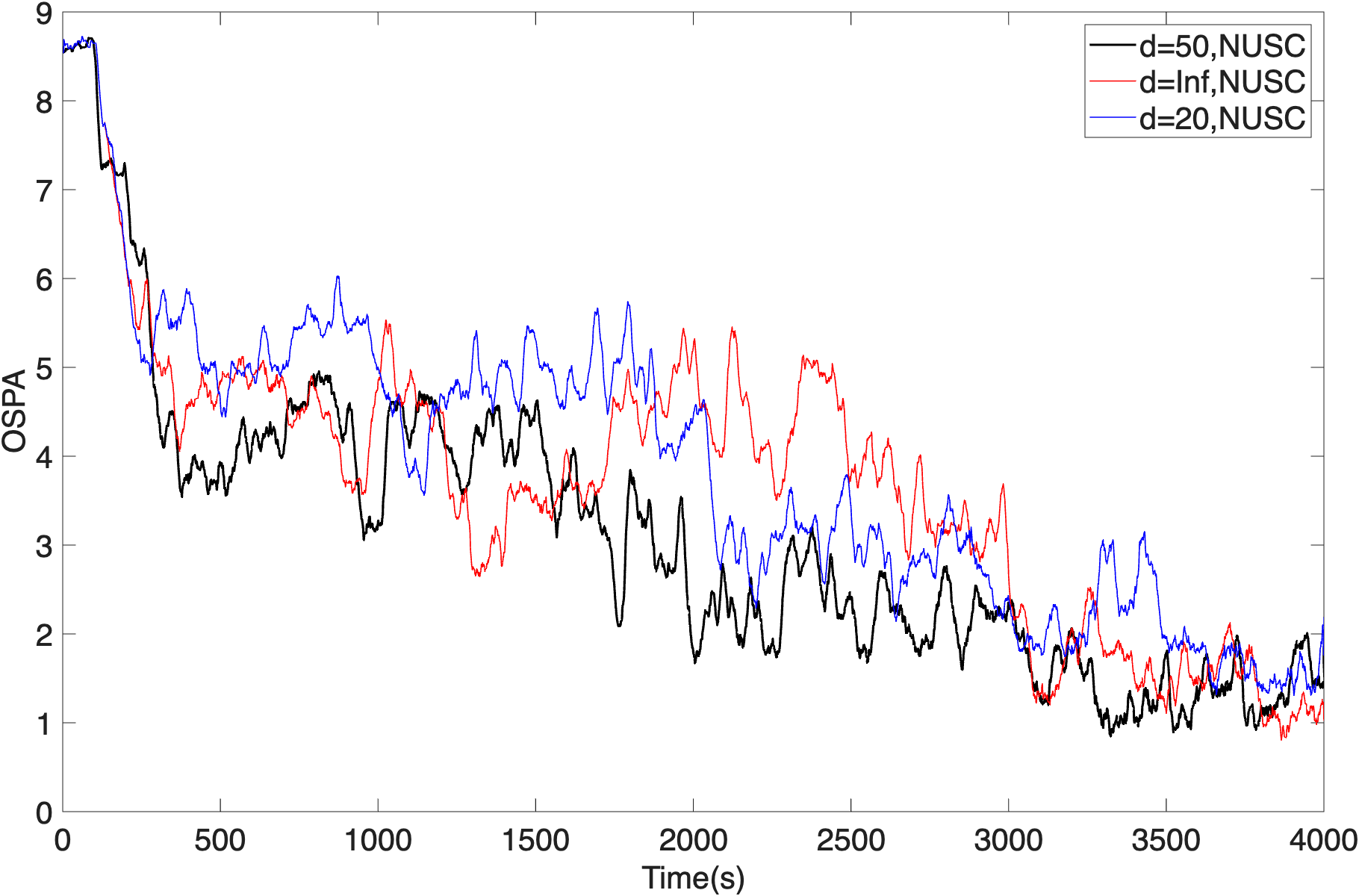}
} \\
\label{fig:d_comparison}
\subfloat[Strategy Comparison]{
\includegraphics[width=0.64\columnwidth]{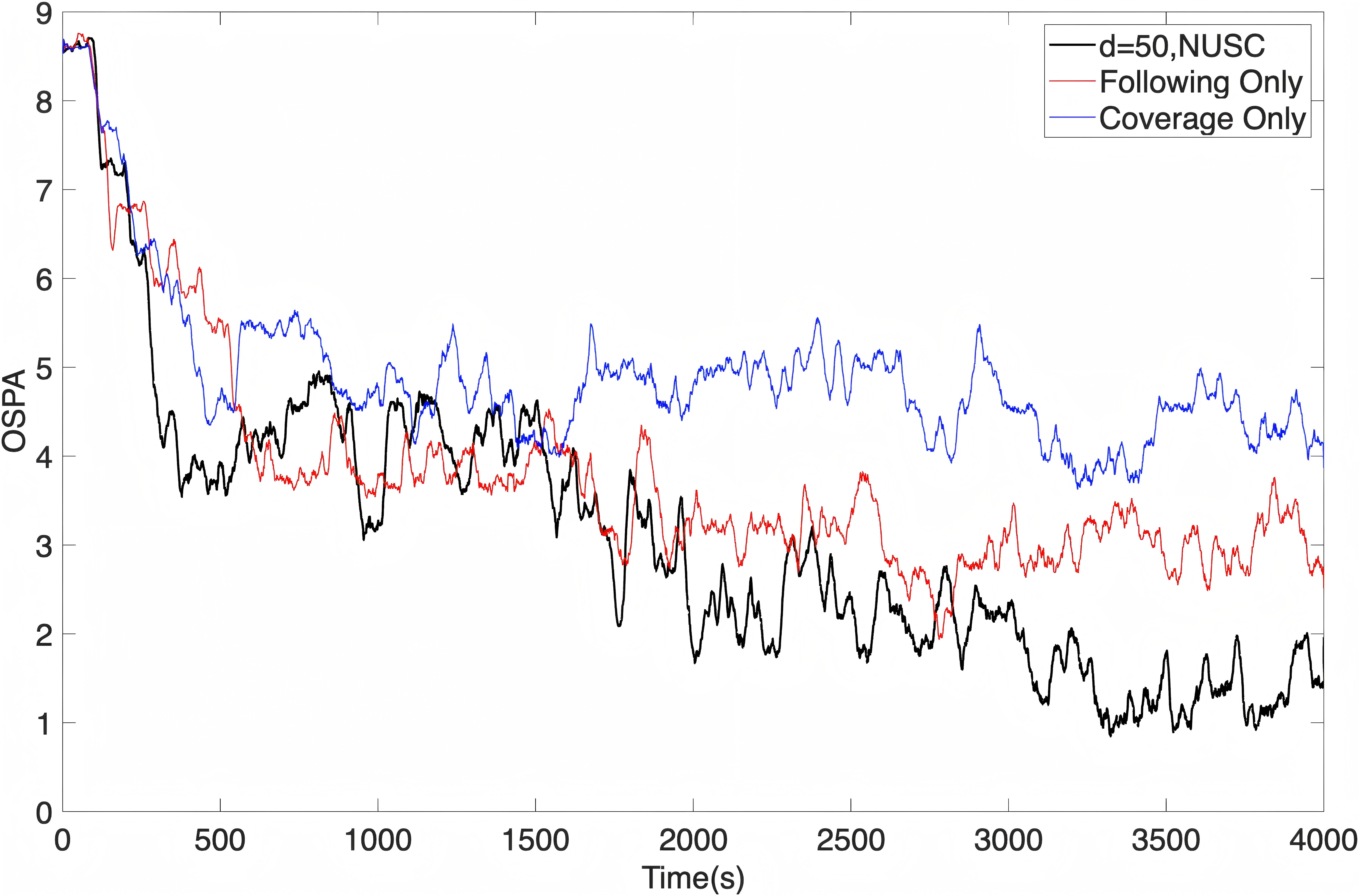}
\label{fig:strategy_comparison}
}
\caption{The figure illustrates the variation in average OSPA values (each over 5 trials) over a 4000-second task duration when employing different algorithms, with a 20-second window used for smoothing. The black curve in all three sub-figures illustrates the method described in this paper, where $d_f$ is set to 50m and the NUSC-weighted power diagram is employed. The colored curves show the OSPA values of compared approaches, respectively.}
\label{fig:error_curve}
\end{figure}
%%%%%%%%%%%%%%%%%%%%%%%%%%%%%%

We conduct simulation experiments using \textsc{MATLAB}.
A team of 20 robots are assigned to explore a $100\textrm{m} \times 100\textrm{m}$ connected indoor environment with 6 rooms and a corridor at a maximum velocity of 2m/s, search for and track 20 targets that move at random velocities less than 1.5m/s towards random directions.
Each robot is equipped with an isotropic sensor with a detection range of 12m and whose field of view is limited by obstacles (e.g., a lidar).
To account for the measurement error, the measurement covariance of a target is set to 0.25m in $g(z \mid x)$ and a 20\% chance of false negative is applied, i.e., $p_d(x) = 0.8, \, \forall x in F_i$.
We set $d_f=50\textrm{m}$, which is approximately twice the diameter of the circular sensing area of a sensor.

\subsection{Qualitative Results}
Figure~\ref{fig:traj} illustrates the complete process of simultaneous search and tracking performed by robots during an experiment.
Initially, the robots start exploring from the bottom area of the corridor. 
In the initial phase, since the area explored by the robots is small, all the robots prioritize coverage search as their primary objective. 
Therefore, they either navigate toward regions with high frontier density or toward the mean centroids of the power cells, as shown in Figures~\ref{fig:500p} and~\ref{fig:500}.
As the exploration area expands, robots originally clustered in the two rooms at the bottom of the map gradually disperse upward to assist robots in the upper regions by exploring uncharted space and conducting coverage searches in newly explored areas, illustrated in Figures~\ref{fig:1000p} and~\ref{fig:1000}.
This demonstrates a form of swarm intelligence, in which robot swarms spontaneously optimize task allocation and autonomously maximize overall target search and tracking performance, given no prior knowledge of the environment.
As the exploration area expands further, robots farther from the frontier begin to navigate toward the PHD-weighted centroid of the power cells to follow targets, thus improving the precision of target tracking, shown in Figures~\ref{fig:1500p},~\ref{fig:2000p}, and~\ref{fig:1500}~\ref{fig:2000}.
Once the entire space has been fully explored, all robots transition into the target-following phase to maximize the accuracy of multi-target tracking, demonstrated by Figures~\ref{fig:3000p},~\ref{fig:4000p}, and~\ref{fig:3000}~\ref{fig:4000}.

During the experiment, we observed an interesting phenomenon. 
Taking the room in the middle of the left and right sides as an example, the robot would repeatedly traverse between the two inner corners of the room, as shown in Figure~\ref{fig:1000}. 
This occurs because during exploration, the highest frontier density alternates between these two corners. 
This behavior enables the robot to achieve better coverage and search efficiency, but simultaneously prolongs the time required to cover the entire space. 
Therefore, when implementing this algorithm in practice, a balance between more thorough coverage search and faster global exploration can be achieved by delaying the execution of robot's moving to it temporary goal that triggers large-scale sudden changes.

\subsection{Quantitative Analysis}

To assess the tracking performance, we use the first-order optimal subpattern assignment (OSPA) metric \cite{4567674}, a widely adopted metric to evaluate the performance of MTT approaches.
Given sets $X$ and $Y$ (representing the true and estimated target locations), the tracking error is defined as
\begin{multline}
d_e(X,Y) = \\ \left( \frac{1}{k} \min_{\pi \in \Pi_{k}} \left( \sum_{i = 1}^{m} d_c(x_i, y_{\pi(i)})^p + c^p (k - m) \right) \right)^{1/p}.
\label{eq:ospa}
\end{multline}
The constant $c$ is a cutoff distance, $d_c(x,y) = \min(c, \|x - y\|)$, and $\Pi_k$ is the collection of all permutations of the set $\{1, 2, \ldots, k\}$.
The higher the value of $p$, the more outliers are penalized.
Without loss of generality, we assume that $|X| = m \leq |Y| = k$ holds. 
In other words, $X$ represents either the true or estimated target set, whichever is smaller.
Equation \eqref{eq:ospa} computes the average matching error between the true and estimated target locations considering all possible assignments between elements $x \in X$ and $y \in Y$ that are within distance $c$.
Note that the \emph{lower} the OSPA value, the more accurate the tracking of the targets.
For each experiment setting, we conduct 5 trials and then take the average OSPA.

\subsubsection{Efficacy of NUSC}
We compare the method employed in this paper—a spatial allocation method using a power diagram weighted by NUSC (plot in black)—with a method that uses a standard Voronoi diagram without considering the current usage of the robot sensing capabilities (plot in red).
In Figure~\ref{fig:voronoi_comparison}, we plot the OSPA values throughout the entire 4000-second simulation. 
It can be seen that both methods enable the OSPA error to converge to a small value, with our method yielding even a smaller one around 0.1, demonstrating the effectiveness of the multi-target simultaneous search and tracking approach from a global perspective.
Both methods exhibit a rapid decrease in OSPA error during the initial 300 seconds. 
This is because the PHD filter updates the target state within the robots' fields of view during the initial phase. 
As the initial update phase concludes, the decline in OSPA begins to slow down.
However, the method proposed in this paper enables the OSPA error to decrease continuously until convergence, whereas the standard Voronoi diagram approach only shows a significant decline in the OSPA error after it fluctuates to nearly 3000 seconds after the complete map is explored.
This demonstrates that after optimizing the operational space through power diagrams and NUSC, the team can more effectively deploy robots to cover and search targets in newly explored areas, thus discovering and tracking more targets at a faster pace.
Note that due to false alarms and errors in the sensors, the OSPA error will not be zero but will stabilize at a small value.

\subsubsection{Testing Various $d_f$}
Then, we set $d_f$ to infinity and 20\textrm{m} in our algorithm for comparison, plotted in blue and green, respectively, in Figure~\ref{fig:d_comparison}, compared to our method, plotted in black.
Both comparison methods can converge the OSPAs to below 0.2 in 4000 seconds.
However, the convergence rates of these two algorithms were slower than when $d_f=50\textrm{m}$, confirming that a moderate value of $d_f$ visibly impacts the speed at which the robot searches for its target.

\subsubsection{Comparison with Single Strategy}
Lastly, we conduct further ablation study to demonstrate the efficacy of our three-strategy structure comparing with using a single strategy.
Figure~\ref{fig:strategy_comparison} plots the results of our method comparing the approach of using Strategy 2 for coverage (plot in blue) and Strategy 3 for target following (plot in red).
Both comparison methods converge to errors significantly higher than our approach. 
Among these, the error incurred by the pure following strategy proves to be the highest, owing to its susceptibility to trapping the robot in local minima within unknown and non-convex environments. 
The coverage-only method also struggles to enable the robot to dynamically adjust its position effectively to maximize tracking accuracy. 
% Consequently, our method, which tightly couples exploration, coverage, and tracking, holds considerable significance.

\section{Conclusions}
We present a novel distributed multi-robot framework for simultaneous environmental exploration and multi-target search and tracking in unknown, non-convex environments. 
By integrating frontier-based exploration with a normalized unused sensing capacity (NUSC)-weighted power diagram and geodesic distance constraints, the proposed approach enables adaptive spatial allocation that effectively balances exploration efficiency with tracking precision. 
The three core strategies -- frontier density-based exploration using kernel density estimation, NUSC-weighted coverage control, and PHD filter-guided target following -- seamlessly coordinate based on the presence of frontiers and a distance threshold, allowing the system to adaptively couples exploration, coverage, and tracking modes. 
Extensive MATLAB simulations demonstrate the capability of the framework to achieve complete coverage exploration while accurately tracking, yielding a faster-converging OSPA error compared to other candidates.

\bibliographystyle{IEEEtran}
\bibliography{references}

\begin{thebibliography}{10}
\providecommand{\url}[1]{#1}
\csname url@rmstyle\endcsname
\providecommand{\newblock}{\relax}
\providecommand{\bibinfo}[2]{#2}
\providecommand\BIBentrySTDinterwordspacing{\spaceskip=0pt\relax}
\providecommand\BIBentryALTinterwordstretchfactor{4}
\providecommand\BIBentryALTinterwordspacing{\spaceskip=\fontdimen2\font plus
\BIBentryALTinterwordstretchfactor\fontdimen3\font minus \fontdimen4\font\relax}
\providecommand\BIBforeignlanguage[2]{{%
\expandafter\ifx\csname l@#1\endcsname\relax
\typeout{** WARNING: IEEEtran.bst: No hyphenation pattern has been}%
\typeout{** loaded for the language `#1'. Using the pattern for}%
\typeout{** the default language instead.}%
\else
\language=\csname l@#1\endcsname
\fi
#2}}

\bibitem{couceiro2011novel}
M.~S. Couceiro, R.~P. Rocha, and N.~M. Ferreira, ``A novel multi-robot exploration approach based on particle swarm optimization algorithms,'' in \emph{2011 IEEE International Symposium on Safety, Security, and Rescue Robotics}.\hskip 1em plus 0.5em minus 0.4em\relax IEEE, 2011, pp. 327--332.

\bibitem{gonzalez2002navigation}
H.~H. Gonz{\'a}lez-Banos and J.-C. Latombe, ``Navigation strategies for exploring indoor environments,'' \emph{The International Journal of Robotics Research}, vol.~21, no. 10-11, pp. 829--848, 2002.

\bibitem{yu2023asynchronous}
C.~Yu, X.~Yang, J.~Gao, J.~Chen, Y.~Li, J.~Liu, Y.~Xiang, R.~Huang, H.~Yang, Y.~Wu, \emph{et~al.}, ``Asynchronous multi-agent reinforcement learning for efficient real-time multi-robot cooperative exploration,'' \emph{arXiv preprint arXiv:2301.03398}, 2023.

\bibitem{yamauchi1997frontier}
B.~Yamauchi, ``A frontier-based approach for autonomous exploration,'' in \emph{Proceedings 1997 IEEE International Symposium on Computational Intelligence in Robotics and Automation CIRA'97.'Towards New Computational Principles for Robotics and Automation'}.\hskip 1em plus 0.5em minus 0.4em\relax IEEE, 1997, pp. 146--151.

\bibitem{burgard2005coordinated}
W.~Burgard, M.~Moors, C.~Stachniss, and F.~E. Schneider, ``Coordinated multi-robot exploration,'' \emph{IEEE Transactions on robotics}, vol.~21, no.~3, pp. 376--386, 2005.

\bibitem{wang2025multi}
C.~Wang, C.~Yu, X.~Xu, Y.~Gao, X.~Yang, W.~Tang, S.~Yu, Y.~Chen, F.~Gao, Z.~Jian, \emph{et~al.}, ``Multi-robot system for cooperative exploration in unknown environments: A survey,'' \emph{arXiv preprint arXiv:2503.07278}, 2025.

\bibitem{renzaglia2020common}
A.~Renzaglia, J.~Dibangoye, V.~Le~Doze, and O.~Simonin, ``A common optimization framework for multi-robot exploration and coverage in 3d environments,'' \emph{Journal of Intelligent \& Robotic Systems}, vol. 100, no.~3, pp. 1453--1468, 2020.

\bibitem{du2021multi}
B.~Du, K.~Qian, H.~Iqbal, C.~Claudel, and D.~Sun, ``Multi-robot dynamical source seeking in unknown environments,'' in \emph{2021 IEEE International Conference on Robotics and Automation (ICRA)}.\hskip 1em plus 0.5em minus 0.4em\relax IEEE, 2021, pp. 9036--9042.

\bibitem{tang2025large}
J.~Tang, Z.~Mao, and H.~Ma, ``Large-scale multi-robot coverage path planning on grids with path deconfliction,'' \emph{IEEE Transactions on Robotics}, 2025.

\bibitem{nair2020mr}
V.~G. Nair and K.~Guruprasad, ``Mr-simexcoverage: Multi-robot simultaneous exploration and coverage,'' \emph{Computers \& Electrical Engineering}, vol.~85, p. 106680, 2020.

\bibitem{senthilkumar2012multi}
K.~Senthilkumar and K.~K. Bharadwaj, ``Multi-robot exploration and terrain coverage in an unknown environment,'' \emph{Robotics and Autonomous Systems}, vol.~60, no.~1, pp. 123--132, 2012.

\bibitem{dames2017detecting}
P.~Dames, P.~Tokekar, and V.~Kumar, ``Detecting, localizing, and tracking an unknown number of moving targets using a team of mobile robots,'' \emph{The International Journal of Robotics Research}, vol.~36, no. 13-14, pp. 1540--1553, 2017.

\bibitem{kantaros2021sampling}
Y.~Kantaros, B.~Schlotfeldt, N.~Atanasov, and G.~J. Pappas, ``Sampling-based planning for non-myopic multi-robot information gathering,'' \emph{Autonomous Robots}, vol.~45, no.~7, pp. 1029--1046, 2021.

\bibitem{chen2025distributed}
J.~Chen, M.~Abugurain, P.~Dames, and S.~Park, ``Distributed multi-robot multi-target tracking using heterogeneous limited-range sensors,'' \emph{IEEE Transactions on Robotics}, 2025.

\bibitem{ramachandran2023resilient}
R.~K. Ramachandran, N.~Fronda, J.~A. Preiss, Z.~Dai, and G.~S. Sukhatme, ``Resilient multi-robot multi-target tracking,'' \emph{IEEE Transactions on Automation Science and Engineering}, vol.~21, no.~3, pp. 4311--4327, 2023.

\bibitem{tolstaya2021multi}
E.~Tolstaya, J.~Paulos, V.~Kumar, and A.~Ribeiro, ``Multi-robot coverage and exploration using spatial graph neural networks,'' in \emph{2021 IEEE/RSJ International Conference on Intelligent Robots and Systems (IROS)}.\hskip 1em plus 0.5em minus 0.4em\relax IEEE, 2021, pp. 8944--8950.

\bibitem{dames2020distributed}
P.~M. Dames, ``Distributed multi-target search and tracking using the phd filter,'' \emph{Autonomous robots}, vol.~44, no.~3, pp. 673--689, 2020.

\bibitem{cortes2004coverage}
J.~Cortes, S.~Martinez, T.~Karatas, and F.~Bullo, ``Coverage control for mobile sensing networks,'' \emph{IEEE Transactions on robotics and Automation}, vol.~20, no.~2, pp. 243--255, 2004.

\bibitem{haumann2013discoverage}
D.~Haumann, V.~Willert, and K.~D. Listmann, ``Discoverage: from coverage to distributed multi-robot exploration,'' \emph{IFAC Proceedings Volumes}, vol.~46, no.~27, pp. 328--335, 2013.

\bibitem{bhattacharya2014multi}
S.~Bhattacharya, R.~Ghrist, and V.~Kumar, ``Multi-robot coverage and exploration on riemannian manifolds with boundaries,'' \emph{The International Journal of Robotics Research}, vol.~33, no.~1, pp. 113--137, 2014.

\bibitem{lajoie2020door}
P.-Y. Lajoie, B.~Ramtoula, Y.~Chang, L.~Carlone, and G.~Beltrame, ``Door-slam: Distributed, online, and outlier resilient slam for robotic teams,'' \emph{IEEE Robotics and Automation Letters}, vol.~5, no.~2, pp. 1656--1663, 2020.

\bibitem{tian2022kimera}
Y.~Tian, Y.~Chang, F.~H. Arias, C.~Nieto-Granda, J.~P. How, and L.~Carlone, ``Kimera-multi: Robust, distributed, dense metric-semantic slam for multi-robot systems,'' \emph{IEEE Transactions on Robotics}, vol.~38, no.~4, 2022.

\bibitem{hougardy2010floyd}
S.~Hougardy, ``The floyd--warshall algorithm on graphs with negative cycles,'' \emph{Information Processing Letters}, vol. 110, no. 8-9, pp. 279--281, 2010.

\bibitem{4567674}
D.~Schuhmacher, B.-T. Vo, and B.-N. Vo, ``A consistent metric for performance evaluation of multi-object filters,'' \emph{IEEE Transactions on Signal Processing}, vol.~56, no.~8, pp. 3447--3457, 2008.

\end{thebibliography}

\end{document}